\pdfoutput=1

\documentclass[11pt]{article}

\usepackage[review]{acl}

\usepackage{graphicx}

\usepackage{times}
\usepackage{latexsym}

\usepackage[T1]{fontenc}

\usepackage[utf8]{inputenc}

\usepackage{microtype}

\title{Geo-located Aspect Based Sentiment Analysis (ABSA) for Crowdsourced Evaluation of Urban Environments}

\author{Demircan Tas \\
  Massachusetts Institute of Technology \\
  \texttt{tasd@mit.edu} \\\And
  Rohit Priyadarshi Sanatani \\
  Massachusetts Institute of Technology \\
  \texttt{sanatani@mit.edu} \\}

\begin{document}
\maketitle
\begin{abstract}
Sentiment analysis methods are rapidly being adopted by the field of Urban Design and Planning, for the crowdsourced evaluation of urban environments. However, most models used within this domain are able to identify positive or negative sentiment associated with a textual appraisal as a whole, without inferring information about specific urban aspects contained within it, or the sentiment associated with them. While Aspect Based Sentiment Analysis (ABSA) is becoming increasingly popular, most existing ABSA models are trained on non-urban themes such as restaurants, electronics, consumer goods and the like. This body of research develops an ABSA model capable of extracting urban aspects contained within geo-located textual urban appraisals, along with corresponding aspect sentiment classification. We annotate a dataset of 2500 crowdsourced reviews of public parks, and train a Bidirectional Encoder Representations from Transformers (BERT) model with Local Context Focus (LCF) on this data. Our model achieves significant improvement in prediction accuracy on urban reviews, for both Aspect Term Extraction (ATE) and Aspect Sentiment Classification (ASC) tasks. For demonstrative analysis, positive and negative urban aspects across Boston are spatially visualized. We hope that this model is useful for designers and planners for fine-grained urban sentiment evaluation.  
\end{abstract}

\section{Introduction}

The field of urban design and planning has been increasingly reliant on sentiment analysis methods for crowdsourced evaluation of urban environments. Point of Interest (POI) data, social media data \cite{marti2019social} and crowd-sourced platforms of citizen engagement \cite{huang2016using} have become powerful tools for gathering large scale descriptive appraisals of the urban realm. Existing NLP methods allow for the extraction of overall valanced sentiment \cite{saeidi2016sentihood} emerging from geo-located textual descriptions. Moreover, unsupervised topic modeling approaches (using methods such as Latent Dirichlet Allocation) allow for the clustering of high-level concepts and topics associated with such descriptions \cite{hu2014social}. However, most models used within this domain are trained to identify positive or negative sentiment associated with a piece of text as a whole. There have been very few attempts at training models that are able to identify the different aspects (semantic features of interest) within a body of text, and the sentiment/opinion associated with each aspect.

This body of research aims to develop an Aspect Based Sentiment Analysis (ABSA) model capable of extracting high level topics/aspects contained within geo-located textual urban appraisals, along with aspect sentiment classification. Such a model will be of immense potential to urban administrators, planners and policy makers, by enabling them in gathering nuanced insights into the everyday aspects of cities that correlate with citizen satisfaction or dissatisfaction. Geo-located aspect-based sentiment analysis also allows for the visualization of the spatial distribution of both aspects as well as sentiment within a city. This also allows for location-based neighborhood evaluation, thus allowing administrators to make decisions based upon knowledge of hyperlocal vulnerabilities.

\section{Related Work}

Aspect Based Sentiment Analysis (ABSA) is slowly emerging as a mainstream research direction \cite{zhang2022survey}. ABSA provides sentiment analysis at the aspect level, unlike more traditional methods of sentiment analysis that operate on sentence or document level. An aspect can be any entity with qualities defined under a certain context. Most ABSA workflows rely upon 4 important categories of semantic content within a document, namely the Aspect Category, Aspect Term, the Opinion Term and the Sentiment Polarity. The aspect term identifies a particular theme/topic/focus of discussion. The opinion term identifies the qualitative opinion associated with that aspect. Finally, the sentiment polarity assigns a sentiment class (such as positive, negative or neutral) to each aspect-opinion pair \cite{zhang2022survey}

Existing word embedding models like Word2Vec \cite{mikolov2013efficient}, and GloVe \cite{pennington2014glove}, while being effective for ABSA, lack the capacity to provide state of the art results. PLM’s provide improved generalization and robustness. However, ABSA tasks require the identification of relationships between tokens, in addition to the identification of token-level, or sequence-level labels.
Pre-trained language models like Bidirectional Encoder Representations from Transformers (BERT) \cite{devlin2018bert}, BART, and RoBERTa \cite{liu2019roberta} have been utilized to overcome the shortcomings of Pre Trained Language Models (PLMs) \cite{qiu2020pre}, resulting in better performance than the previous state of the art in solving ABSA tasks. The self-attention mechanism of PLMs captures full word dependencies in a sentence, which is redundant in terms of ABSA tasks \cite{nazir2020issues}. While there is still room for improvement in adapting PLMs for effective use in ABSA tasks, especially in terms of robustness, they provide significant improvements over previous methods, making them the common answer in the NLP community for ABSA tasks.

While ABSA is becoming increasingly popular, most existing ABSA datasets are currently focused on non-urban themes such as restaurants \cite{jiang2019challenge}, electronics \cite{nakov2019semeval}, consumer goods \cite{wang2019aspect} and the like. Models trained in such datasets are not relevant to ABSA tasks within the urban domain. The ability to identify opinions and sentiment associated with specific aspects of urban experience will be useful for spatial analysts working with textual data. 

\section{Methodology}
\subsection{Data Source}
While numerous platforms exist today that can serve as valuable repositories of textual descriptions of urban environments, the chosen platform needed to fulfill some basic criteria. Firstly, the data needed to be of high spatial resolution. In other words, the data should be geo-located with an accuracy that is high enough for meaningful inferences with respect to its location to emerge. Secondly, the data should also be of high temporal resolution. In other words, the exact time and date of the review/photographs should also be available. Finally, the data should contain adequate and relevant semantic content with regards to the different qualitative aspects of urban experience, which is often a challenge.
Based on these criteria, Point of Interest (POI) data sources emerged as the most appropriate in this regard. While there are numerous major POI data aggregators globally (such as Google Places, Foursquare and Yelp) \cite{niu2020crowdsourced}, this body of research adopted Google Places as the primary data source for analysis. 
Data for points of interest listed as ‘parks’ was programmatically collected across the greater Boston area, using the Google Places API. The first round of data collection involved the collection of basic place details using the ‘Nearby Search’ API call. Query locations and radii were set at regular intervals across the cities, and places data queried from each location. The data included attributes such as location, place name, place id, rating, total user ratings and the like. Next, for each of the places now listed in the dataset, 5 top reviews were programmatically collected using the ‘Place Details’ API call. For each review, the query returned attributes such as a review text, author name, language, rating, and a timestamp. In total, 2500 reviews were collected. Highlighted below are some representative examples from our dataset. 

\begin{quote}
    \textit{"Well it's not the most pleasant conservation area, but they do try. The trails are well maintained and groomed. But in spite of that the conservation area itself has lots of trash in it, it's overgrown with nasty and prickly plants, and it's muddy. Not the best place to take a walk in nature..."}

    \textit{"A wonderful place, you find yourself in a fairy tale. A very interesting and uncomplicated trail, along the way there are many places for a wonderful picnic in nature. The wonderful smell and singing of birds will accompany you all the way."}
\end{quote}

\subsection{Data Labelling}
Aspect terms and corresponding sentiment labels (‘Positive’, ‘Negative’,’Neutral’) were manually identified and coded into the dataset for evaluation. This was done through the manual dataset annotation tool developed by \cite{yang2022pyabsa}. Moreover, the overall sentiment of each review was also manually labelled for conventional sentiment analysis tasks.

\subsection{Model selection}
A Bidirectional Encoder Representations from Transformers (BERT) \cite{devlin2018bert} with Local Context Focus as outlined by \cite{yang2021multi} was chosen for ABSA training. This was based on a critical review of existing literature and evaluation of baseline results. Most importantly such a model is well suited for concurrent Aspect Term Extraction (ATE) and Aspect Polarity Classification (APC) in a single forward pass. Figure 1 below describes the model architecture. 

$\vcenter{\hbox{\includegraphics[width = 75mm]{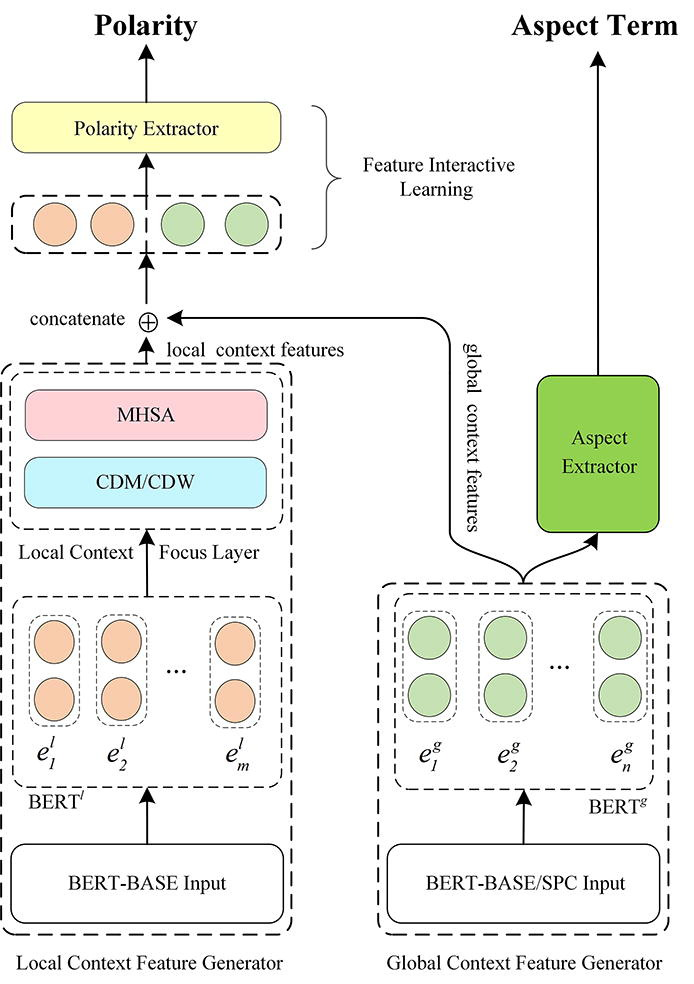}}}$ 

Fig 1: ABSA Model. Source: \cite{yang2021multi}

\subsection{Dataset Generation}
The annotated data was converted into a .txt file, such that a specific review was repeated in the .txt for every aspect that it contained. The position of the aspect term was denoted using the \textit{\$T\$} marker. The aspect term was identified on the next line, followed by the aspect polarity label assigned to it. Shown below is a representative sample of the output format for a single review. 

\begin{quote}
Nice \$T\$ for sport activities the area is very clean

playground

Positive

Nice playground for \$T\$ activities the area is very clean

sport

Positive

Nice playground for sport activities the \$T\$ is very clean

area

Positive

\end{quote}

This textual input was then converted into a format that could directly be fed into the BERT model for final vectorisation. This was done using the PyABSA library \cite{yang2022pyabsa}. Shown below is dataset sample in its final form. 

\begin{quote}
    Nice O -999
playground B-ASP Positive
for O -999
sport B-ASP -999
activities O -999
the O -999
area B-ASP -999
is O -999
very O -999
clean O -999

Nice O -999
playground B-ASP -999
for O -999
sport B-ASP Positive
activities O -999
the O -999
area B-ASP -999
is O -999
very O -999
clean O -999

Nice O -999
playground B-ASP -999
for O -999
sport B-ASP -999
activities O -999
the O -999
area B-ASP Positive
is O -999
very O -999
clean O -999
\end{quote}

\subsection{Model training}
Table 1. below describes the key training parameters. 

 \begin{table}[h]
     \centering
     \begin{tabular}{c|c|c|c}
         Training size & 2250 & Test size & 250  \\
              \hline
          Batch size & 16 & num epochs & 6 \\
     \end{tabular}
     \caption{Key training parameters}
     \label{tab:my_label}
 \end{table}
 $\vcenter{\hbox{\includegraphics[width = 75mm]{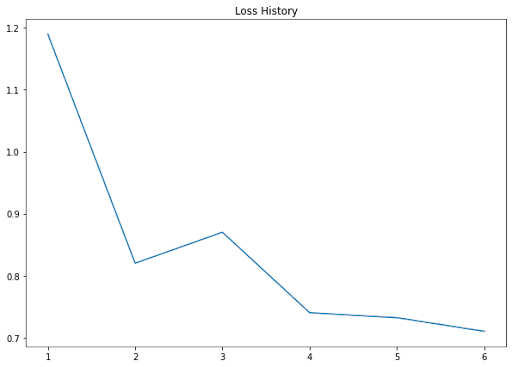}}}$ 

\section{Results} 
\subsection{Model performance}

Our model performs significantly better than other baseline models tested out as part of our methodology. We report improved accuracy for both Aspect Term Extraction (ATE) as well as Aspect Polarity Classification (APC) tasks. This was the aim of this work, as previous models tested out were trained on non-urban datasets and thus were unable to identify many aspects from the input text. Table 2 below describes a comparative evaluation of F1 scores for a baseline pre-trained BERT based model and our model trained on our urban review dataset. 

 \begin{table}[h]
     \centering
     \begin{tabular}{c|c|c}
          & ATE & APC  \\
              \hline
          Baseline (Pretrained BERT) & 0.7921 & 0.7023 \\
          Our Model (LCF-ATEPC) & 0.83321 & 0.7476 \\ 
     \end{tabular}
     \caption{Evaluation Metrics}
     \label{tab:my_label}
 \end{table}

Qualitative evaluation revealed improved performance as well. In the baseline model, many tokens which were legitimate urban aspects were missed out and extracted as part of the ATE workflow. Our model missed fewer aspects because of training on an urban dataset, and thus reported higher ATE scores. Shown below is a representative example of model predictions. 

\begin{quote}
    Review:  A wonderful place , you find yourself in a fairy tale . A very interesting and uncomplicated trail , along the way there are many places for a wonderful picnic in nature . The wonderful smell and singing of birds will accompany you all the way .
    
Aspect:  ['place', 'trail', 'picnic', 'birds']

Sentiment:  ['Positive', 'Positive', 'Positive', 'Positive']

\end{quote}

\subsection{Demonstrative Urban Sentiment Evaluation}
We present a demonstrative urban sentiment evaluation of urban public parks in Boston, to highlight the impact of this model within the field of urban planning and analytics.
\subsection{Urban Aspect Analysis}
It is often necessary to identify the key aspects associated with with both positive and negative sentiment, so as to inform policy considerations or design/planning interventions with respect to those aspects. This allows for fine grained aspect based sentiment evaluations of the public realm, which goes beyond broad brush sentiment classification of places which are common today. 

The figures below illustrate the frequency distributions of the top positive and negative aspects as analysed through sentiment extraction of crowd-sourced reviews through our model.

$\vcenter{\hbox{\includegraphics[width = 75mm]{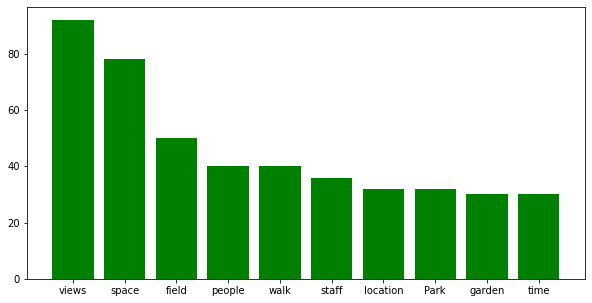}}}$

$\vcenter{\hbox{\includegraphics[width = 75mm]{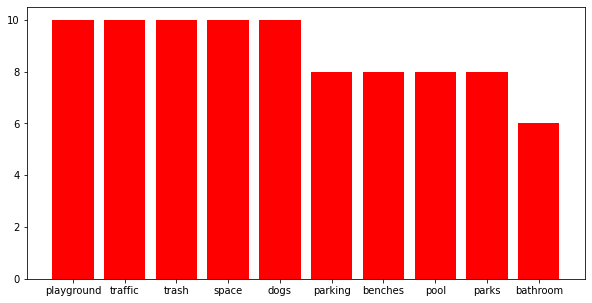}}}$

Fig 2: Frequency distribution of key positive and negative urban aspects

\subsection{Geo-spatial aspect analysis}

For urban planning and design, it is important to not only evaluate positive and negative aspects, but also the geographical distribution of aspects. Since most urban reviews collected from urban POI aggregators are geo-located, fine grained aspect based spatial analysis is possible using our model. Shown below are few demonstrative examples of the spatial distribution of urban aspects classified by our model. 

\hbox{\includegraphics[width = 75mm]{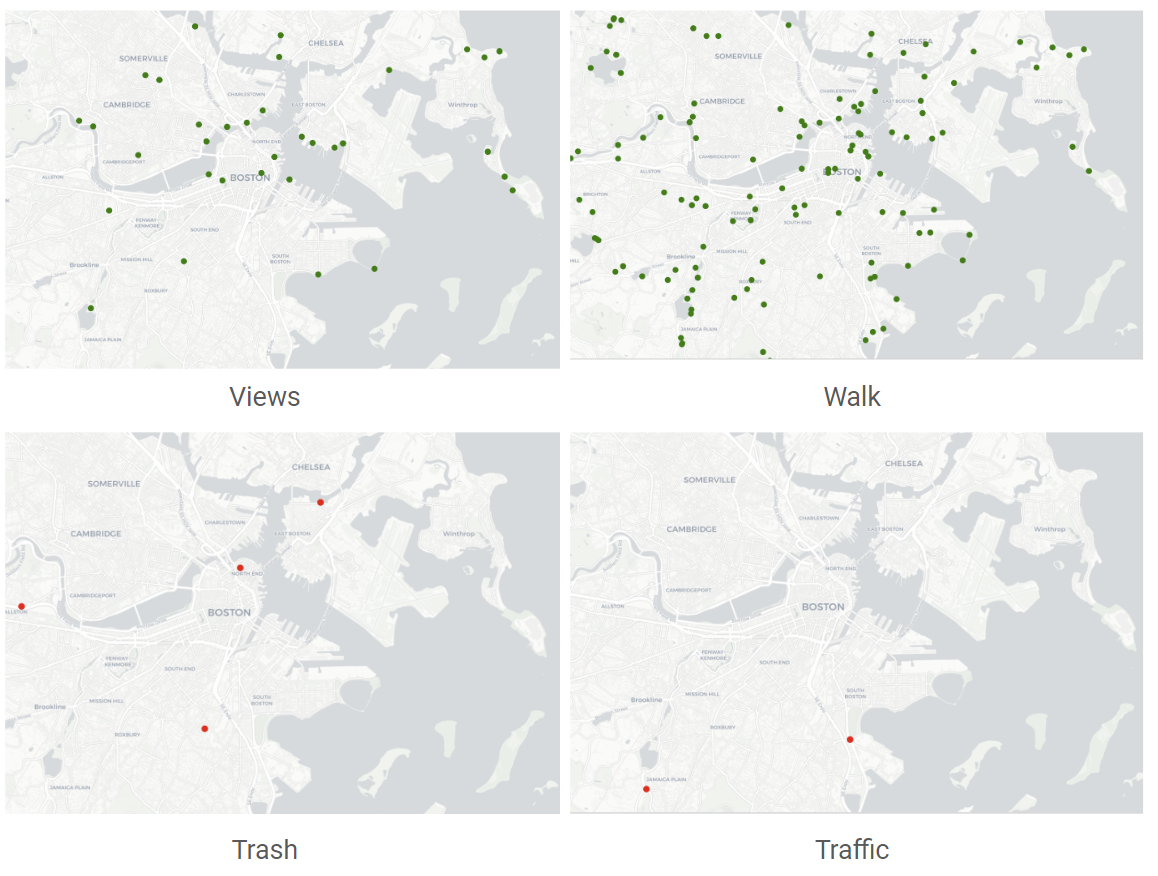}}
Fig 3: Spatial distribution of positive and negative urban aspects in Boston

\section{Conclusion - Impact}

This body of research aimed to develop an Aspect Based Sentiment Analysis (ABSA) model that was capable of fine grained urban aspect identification and sentiment classification based on crowd-sourced appraisals of urban environments. We see from the performance of our model, and from the demonstrative analyses on real-world datasets, that our framework succeeds in this task and performs better than existing models trained on non-urban data. ABSA is an emerging research domain, and it is slowly adapted for different domains. To the best of our knowledge, this work is one of the first systematic adaptations of ABSA model architectures for crowdsourced evaluation of urban public spaces. Moreover, we hope that the dataset we generated as part of this work is taken up by future researchers for further experimentation with novel model architectures. We also hope that our framework is integrated into mainstream urban planning workflows, for data-driven decision support and more democratic forms of urban governance. 

\bibliography{custom}
\bibliographystyle{acl_natbib}

\end{document}